\documentclass[letterpaper,10pt,conference]{ieeeconf}

\IEEEoverridecommandlockouts
\usepackage{amsmath}
\usepackage{amssymb}
\usepackage{graphicx}
\usepackage{url}
\usepackage[draft]{hyperref}

\title{\LARGE \bf
	ContactIPM: A Structure-Exploiting Interior-Point Solver for
	Contact-Implicit Trajectory Optimization
}

\author{Yucheng Chen}

\begin{document}
	
	\maketitle
	\thispagestyle{empty}
	\pagestyle{empty}
	
\begin{abstract}
	Contact-implicit trajectory optimization avoids prescribing contact
	sequences, but yields mathematical programs with complementarity constraints
	(MPCCs) whose degeneracy challenges conventional primal--dual solvers.
	Existing contact-specific methods improve robustness to this degeneracy but
	do not leverage a stagewise optimal-control
	factorization and primal--dual consistency, while structure-exploiting optimal-control solvers are not
	designed for complementarity constraints. We show that these capabilities can
	be combined in a single primal--dual method. ContactIPM identifies
	complementary inequality pairs, embeds them through a barrier-coupled elastic
	interior relaxation, eliminates slack and dual variables stagewise, and
	solves the reduced Newton system using a Riccati recursion. A fixed multi-phase MPCC recovery schedule provides four continuation and
	restart attempts from naive initializations, while termination is gated
	by the unrelaxed physical complementarity residual.
	
	We compare ContactIPM with two contact-specific MPCC solvers, CRISP and
	IMPACT, using matched benchmark conditions and common post-solve acceptance
	criteria. On four fixed CRISP benchmark cases, ContactIPM is
	$2.17$--$8.87\times$ faster over 20 paired timing repetitions per case
	and achieves higher success on the Push Box and Push-T robustness
	suites.
	Against IMPACT, ContactIPM is \(2.96\times\) faster on Push T and \(4.91\times\) faster on Cart
	Transport, but \(4.46\times\) slower on Push Box. In 50 closed-loop Push Box
	rollouts spanning model mismatch, measurement noise, initial-pose errors, and
	state resets, ContactIPM succeeds in all 50 cases with a median solve time of
	2.08\,ms. Additionally, an OCP-structured acados baseline imposing exact complementarity
	frequently fails to generate useful contact, supporting the need for
	MPCC-specific treatment in addition to temporal-structure exploitation.
	Together, these results demonstrate that MPCC robustness,
	primal--dual consistency, and optimal-control structure can coexist in
	one solver.
	
	An anonymized implementation and reproduction scripts are available
	online.\footnote{\url{https://anonymous.4open.science/r/ContactIPM-0C58}}
\end{abstract}
	
	\section{Introduction}
	\label{sec:introduction}
	
	Robots that push, slide, grasp, or locomote must reason about both continuous
	motion and discrete changes in contact. Contact-implicit trajectory
	optimization (CITO) avoids fixing a contact schedule by optimizing states,
	inputs, contact forces, and contact modes simultaneously
	\cite{posa2014direct,manchester2017variational,sleiman2021contact}.
	This flexibility comes at a numerical cost: unilateral contact and friction
	are expressed through complementarity, yielding an MPCC that violates the
	classical constraint qualifications used by smooth nonlinear programming.
	
Recent contact-implicit solvers obtain robustness by departing from a fully
coupled primal--dual Newton step. CRISP optimizes only the primal trajectory
through a sequence of convex subproblems with trust-region globalization
\cite{li2025crisp}; subproblem multipliers are not maintained as part of a
single primal--dual iterate for the original MPCC. IMPACT uses a safeguarded
augmented-Lagrangian outer loop and an inner block-coordinate descent (BCD)
scheme that alternates trajectory updates with closed-form complementarity
updates \cite{li2026impact}. IMPACT therefore maintains augmented-Lagrangian
multipliers, but its primal and complementarity blocks are not updated by
solving one fully coupled perturbed KKT system. These choices are effective
responses to MPCC degeneracy, but they separate complementarity handling from
the primal--dual structure exploited by conventional interior-point methods.

We argue that this separation is not an unavoidable cost of contact
complementarity. In a multiple-shooting transcription, each complementarity
pair is local to one stage. Its relaxation, slack variables, multipliers, and
derivative contributions are therefore also stage local. Eliminating these
local slack and dual variables modifies only the corresponding stage gradient
and Hessian; the dynamics remain the only coupling between adjacent stages.
Consequently, complementarity constraints do not destroy the block-banded OCP
structure required by a Riccati recursion.

ContactIPM builds on this observation by introducing a stage-local elastic
interior relaxation for every marked complementarity pair. The construction
has two important consequences. First, for every positive barrier parameter,
the lifted constraint Jacobian contains an identity block with respect to the
elastic slacks. The relaxed complementarity rows therefore satisfy LICQ within
our multiple-shooting formulation, and their multipliers are uniquely defined
at a stationary point, which provides a regular primal--dual system. Second,
the complementarity relaxation is coupled directly to the interior-point
barrier. Tightening is controlled by the same barrier parameter that defines
the central path, rather than by an independently tuned penalty weight that
must be repeatedly increased.

These observations define the capability gap targeted by ContactIPM.
Contact-specific methods such as CRISP and IMPACT obtain MPCC robustness
without retaining a single coupled primal--dual OCP Newton system, whereas
OCP-structured solvers such as acados \cite{verschueren2022acados}  can fail to generate useful contact when
exact complementarity is imposed without special treatment (Section \ref{sec:acados_baseline}). ContactIPM
combines both capabilities: a barrier-coupled elastic path for robust
contact-mode transitions and a stagewise primal--dual Riccati solve. When no complementarity pairs are marked, the implementation leaves the
MPCC-specific machinery inactive and reduces to a
constrained-OCP interior-point solver.
	
\begin{figure*}[t]
	\centering
	
	\begin{minipage}[t]{0.35\textwidth}
		\vspace{0pt}
		\centering
		\includegraphics[width=\linewidth]
		{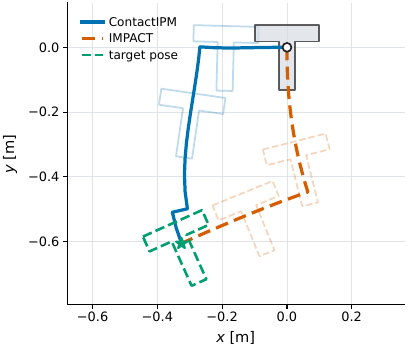}
		\par\vspace{1mm}
		{\small (a) Push-T trajectories}
	\end{minipage}%
	\hfill
	\begin{minipage}[t]{0.32\textwidth}
		\vspace{0pt}
		\centering
		\includegraphics[width=\linewidth]
		{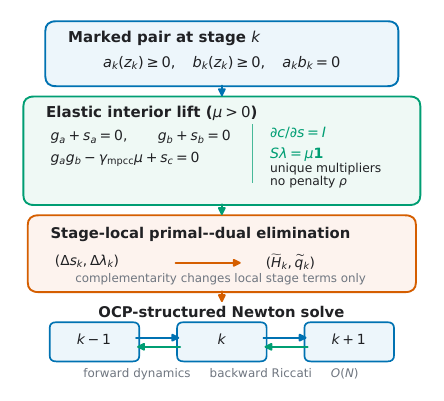}
		\par\vspace{1mm}
		{\small (b) ContactIPM structure}
	\end{minipage}%
	\hfill
	\begin{minipage}[t]{0.31\textwidth}
		\vspace{0pt}
		\centering
		\includegraphics[width=\linewidth]
		{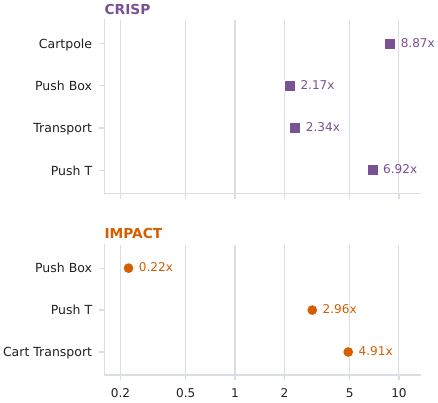}
		\par\vspace{1mm}
		{\small (c) Runtime comparison}
	\end{minipage}
	
	\caption{
		Overview of the proposed solver and representative experimental
		results.
		(a) On a matched Push-T experiment, ContactIPM and IMPACT reach the same target from the
		same initialization but follow different trajectories, illustrating
		convergence to different local solutions.
		(b) ContactIPM introduces elastic variables for the stagewise
		complementarity constraints. The resulting regular constraint
		representation admits consistent primal--dual iterates and
		well-defined multipliers, while local elimination preserves the
		optimal-control structure required by the Riccati recursion.
		(c) Median runtime ratios between each competitor and ContactIPM.
		The dashed vertical line denotes equal runtime; values greater than
		one indicate that ContactIPM is faster.
	}
	\label{fig:overview}
\end{figure*}
	
	The contributions are:
\begin{itemize}
	\item a unified stagewise formulation for nonlinear dynamics, costs,
	inequalities, bounds, and marked complementarity pairs, in which the
	latter are converted into barrier-coupled elastic product rows;
	
	\item an MPCC-aware primal--dual Newton method that preserves the perturbed
	KKT coupling while using local slack and dual elimination followed by a
	Riccati recursion, together with trajectory-scale preconditioning and
	multi-phase MPCC recovery. The elastic lifting and stagewise elimination are summarized in
	Fig.~\ref{fig:overview}(b); and
	
	\item a reproducible experimental evaluation comprising
	CRISP and IMPACT comparisons, an OCP-structured acados baseline with exact
	complementarity, and 50 disturbed closed-loop Push Box rollouts. The representative Push-T trajectories in Fig.~\ref{fig:overview}(a) show
	that ContactIPM and IMPACT converge to different local solutions; and the timing ratios summarized in Fig.~\ref{fig:overview}(c) show
	performance across both baselines.
\end{itemize}
	
	\section{Background and Problem Formulation}
	\label{sec:problem}
	
	\subsection{Contact-Implicit Optimal Control}
	
	We consider a multiple-shooting problem over horizon $N$,
	\begin{subequations}
		\label{eq:ocp}
		\begin{align}
			\min_{\mathbf{x},\mathbf{u}}\quad&
			\sum_{k=0}^{N-1}\ell_k(x_k,u_k)+\ell_N(x_N),
			\label{eq:ocp_cost}\\
			\text{s.t.}\quad&
			x_0=\bar{x}_0,\qquad
			x_{k+1}=f_k(x_k,u_k),                         \label{eq:ocp_dyn}\\
			&   g_k(x_k,u_k)\leq 0,                           \label{eq:ocp_ineq}\\
			&   0\leq a_{k,p}(x_k,u_k)\perp
			b_{k,p}(x_k,u_k)\geq 0,                    \label{eq:ocp_mpcc}
		\end{align}
	\end{subequations}
	where $p$ indexes complementarity pairs. Depending on the contact model,
	$a_{k,p}$ and $b_{k,p}$ may represent normal force and separation, frictional
	force and slip, or mutually exclusive contact modes. Their complementarity is
	equivalent to nonnegativity together with
	$a_{k,p}b_{k,p}=0$.
	
	At a feasible MPCC point, the gradients of the active nonnegativity and
	product constraints are structurally dependent. Standard constraint qualifications therefore fail at feasible MPCC
	points, making direct application of generic nonlinear-programming
	methods unreliable
	\cite{biegler2005mpcc,leyffer2006interior}. This degeneracy is
	particularly problematic for primal--dual methods because the
	associated multipliers may be nonunique and the KKT system may become
	singular or ill-conditioned.
	Relaxation methods restore a nonempty interior by replacing exact
	complementarity with a parameterized inequality, but aggressive tightening can
	again cause ill-conditioning.
	
	\subsection{The Structure--Robustness Gap}
	
	Classical MPCC interior methods provide convergence results for suitably
	relaxed formulations \cite{biegler2005mpcc,leyffer2006interior}. Building on these established MPCC relaxation theory, ContactIPM embeds
	this machinery in a multistage optimal-control solver, retains a direct
	physical-complementarity stopping test, and provides generic recovery when the
	relaxed central path becomes difficult.
	
	CRISP and IMPACT approach the MPCC side of the gap differently. CRISP avoids
	maintaining a coupled primal--dual iterate and solves a sequence of convex
	primal subproblems. IMPACT maintains safeguarded augmented-Lagrangian
	multipliers and performs closed-form updates for complementarity pairs, while
	solving its trajectory subproblems with a sparse factorization. These are
	substantive uses of problem structure, but neither reduces the coupled
	trajectory Newton system with a Riccati recursion.
	
	On the OCP side, Riccati and block-factorization techniques are established
	tools for exploiting horizon structure \cite{domahidi2012efficient}. Hippo recently demonstrated a high-performance primal--dual interior-point and projection-based solver for constrained trajectory optimization \cite{zhao2026hippo}. FilterDDP \cite{xufilterDDP2026} similarly extends structure-exploiting DDP with a primal--dual interior-point treatment of generic inequalities and evaluates the method on contact-implicit tasks, where contact is supplied through task-specific equality and slack-variable encodings. Although both methods have numerical cores related to ContactIPM's, neither provides MPCC-specific pair recognition, barrier-coupled elastic product rows, an unrelaxed physical-complementarity acceptance gate, or an MPCC recovery policy. ContactIPM is positioned at this intersection: it operates as an
	IPM-based constrained-OCP solver when no pairs are marked and activates
	its MPCC-specific machinery only where complementarity is present.
	
	\section{ContactIPM}
	\label{sec:method}
	
	\subsection{MPCC and Elastic Interior Relaxation}
	
	The model supplies ordinary one-sided rows $g_j(x,u)\leq0$ and marks pairs
	$(g_a,g_b)$ whose negations represent complementary nonnegative quantities:
	\begin{equation}
		a=-g_a\geq0,\qquad b=-g_b\geq0,\qquad ab=0.
	\end{equation}
	For each marked pair, ContactIPM introduces positive side slacks $s_a,s_b$
	and a positive product slack $s_c$:
	\begin{subequations}
		\label{eq:elastic}
		\begin{align}
			g_a+s_a &= 0,\qquad s_a>0,\\
			g_b+s_b &= 0,\qquad s_b>0,\\
			g_ag_b-\theta(\mu)+s_c &= 0,\qquad s_c>0,\\
			\theta(\mu)&=\gamma_{\mathrm{mpcc}}\mu ,
		\end{align}
	\end{subequations}
	where $\mu>0$ is the barrier parameter. The barrier objective includes
	\begin{equation}
		-\mu\left(\log s_a+\log s_b+\log s_c\right),
	\end{equation}
	so the primal--dual equations contain
	$s_i\lambda_i=\mu$ for both side rows and the generated product row. The
	formulation is therefore an elastic interior relaxation rather than a
	penalty-only treatment.
	
	The relaxed equation alone is not used to certify a contact solution. At every
	candidate termination point, ContactIPM also evaluates
	\begin{equation}
		r_{\mathrm{phys}}
		=\max_{k,p}\left|a_{k,p}b_{k,p}\right|,
		\label{eq:physical_residual}
	\end{equation}
	and requires $r_{\mathrm{phys}}\leq\epsilon_{\mathrm{mpcc}}$. This separates
	the mechanism used to maintain an interior during optimization from the
	physical condition used to accept the final trajectory.
	
	\paragraph{Regularity of the elastic lift.}
	For a fixed $\mu>0$, and decision variables $z$, collect the three residuals associated with a marked pair
	as
	\begin{equation}
		c_{\mathrm{el}}(z,s)=
		\begin{bmatrix}
			g_a(z)+s_a\\
			g_b(z)+s_b\\
			g_a(z)g_b(z)-\theta(\mu)+s_c
		\end{bmatrix}=0 .
	\end{equation}
	Their Jacobian with respect to the local slack variables is
	\begin{equation}
		\frac{\partial c_{\mathrm{el}}}{\partial(s_a,s_b,s_c)}
		= I_3 .
		\label{eq:elastic_slack_pivot}
	\end{equation}
	Thus, the lifted complementarity rows are linearly independent at every
	interior point, irrespective of the gradients of $g_a$ and $g_b$. Ordinary
	inequalities have the same property after slack lifting. Moreover, each
	multiple-shooting dynamics defect contains an identity block with respect to
	$x_{k+1}$. Consequently, the equality Jacobian of the relaxed problem has full
	row rank for the problem class considered here.
	
This establishes LICQ for every relaxed subproblem solved by ContactIPM.
Therefore, if a stationary point exists, its equality multipliers are unique.
The positive slacks also determine the inequality multipliers uniquely through
\begin{equation}
	S\lambda=\mu\mathbf{1},
	\qquad
	\lambda=\mu S^{-1}\mathbf{1}.
\end{equation}
Because ContactIPM maintains $\mu>0$ and strictly positive slacks throughout
the interior-point iterations, this regularity holds for every Newton system
encountered by the solver.
	
	\paragraph{No independent complementarity penalty weight.}
	Penalty and augmented-Lagrangian approaches typically introduce weights that
	must be increased until complementarity violation is sufficiently expensive.
	In ContactIPM, the generated product row is instead enforced as a primal
	equality, while its positive elastic slack is handled by the logarithmic
	barrier. The target
	\begin{equation}
		\theta(\mu)=\gamma_{\mathrm{mpcc}}\mu
	\end{equation}
	ties complementarity tightening to the same $\mu$ used by the primal--dual
	central path. ContactIPM therefore requires no independently escalated
	complementarity penalty coefficient. The parameter
	$\gamma_{\mathrm{mpcc}}$ and the barrier-update policy remain numerical
	settings, so the method is not parameter free, but it avoids balancing a
	separate penalty magnitude against the original OCP objective.
	
	\subsection{Primal--Dual Newton System}
	
	ContactIPM maintains primal--dual consistency by computing every direction
	from the linearization of one perturbed KKT system:
	\begin{subequations}
		\label{eq:perturbed_kkt}
		\begin{align}
			\nabla_{\mathbf{x},\mathbf{u}}\mathcal{L} &= 0,\\
			x_{k+1}-f_k(x_k,u_k) &= 0,\\
			g_k(x_k,u_k)+s_k &= 0,\\
			S_k\lambda_k-\mu\mathbf{1} &= 0.
		\end{align}
	\end{subequations}
	Here, primal--dual consistency means that the primal trajectory, dynamics
	multipliers, inequality multipliers, and slacks are updated as components of
	the same Newton system. In particular, the multipliers are neither omitted, as in CRISP's
	primal-only formulation, nor updated through a separate block-coordinate
	step, as in IMPACT. ContactIPM therefore retains the coupled
	perturbed-KKT structure required by a primal--dual interior-point method.
	
	Crucially, this coupling does not require a monolithic trajectory
	factorization. Complementarity pairs and their elastic variables are local to
	individual stages. Let $C_k$ denote the Jacobian of the original stage inequalities and
	the product-relaxation rows introduced for the marked complementarity
	pairs. Eliminating $\Delta s_k$ and $\Delta\lambda_k$ from the
	same linearized KKT equations produces
	\begin{align}
		\widetilde{H}_k
		&= H_k+C_k^\top
		\operatorname{diag}\!\left(\lambda_k\slash s_k\right)C_k,\\
		\widetilde{q}_k
		&= q_k+C_k^\top w_k ,
	\end{align}
	where $w_k$ collects the local primal-feasibility and centering residuals.
	Both corrections remain confined to stage $k$. The reduced system retains the block-banded equality-constrained OCP
	structure induced by the dynamics. It is solved by a backward Riccati
	recursion followed by forward substitution, after which the eliminated
	slack and multiplier directions are recovered from the same KKT
	equations. For fixed state and input dimensions, the recursion has
	linear complexity in the horizon length~$N$. This factorization is
	used whether or not marked complementarity constraint pairs are present; complementarity changes the
	local stage terms but not the temporal elimination pattern. Exact Lagrangian
	curvature is used when its stage factorizations are sufficiently positive
	definite; otherwise, the solver falls back to Gauss--Newton curvature.
	
	\subsection{Scaling and Globalization}
	
	Contact-rich problems mix positions, angles, forces, impulses, and product
	rows with substantially different numerical scales. ContactIPM applies a
	horizon-consistent diagonal coordinate scaling derived from stage derivative
	magnitudes. Dynamics, cost, constraints, and Riccati quantities are transformed
	consistently, and the accepted Newton step is mapped back to physical
	coordinates.
	
	Globalization uses a filter line search that balances objective reduction and
	constraint violation \cite{wachter2006implementation}. A fraction-to-boundary
	rule maintains positive slacks and multipliers. A nonlinear dynamics rollout
	is also evaluated as a candidate step so that a direction that is useful for the
	contact variables is not rejected solely because of a multiple-shooting
	defect. The barrier parameter is reduced only after the current subproblem has
	reached the required primal, dual, and complementarity accuracy.
	
\subsection{Multi-Phase MPCC Recovery}
\label{sec:recovery}

The relaxed MPCC may contain several stationary branches associated with
different contact modes. During barrier reduction, a solve can enter an
unfavorable basin, lose conditioning, or encounter unreliable second-order
curvature. ContactIPM therefore combines continuation, which preserves the
current basin, with restart, which searches for a different one.

Let
\begin{align}
	\mu_{\mathrm{t}}
	&= \min(\mu_0,\mu_{\mathrm{conv}}), \\
	\mu_{\mathrm{m}}
	&= \sqrt{\mu_0\mu_{\mathrm{t}}}, \\
	\mu_{\mathrm{r}}
	&= \max(\mu_{\mathrm{recovery}},\mu_{\mathrm{conv}}).
\end{align}
The solver first uses the user-specified barrier and curvature
configuration. If unsuccessful, it retains
the latest primal trajectory and solves successively from
$\mu_{\mathrm{m}}$ and $\mu_{\mathrm{t}}$. These phases gradually tighten
$\theta(\mu)=\gamma_{\mathrm{mpcc}}\mu$ while attempting to remain in the
contact-mode basin found by the primary solve. If continuation fails, the
original primal guess is restored and the problem is restarted at
$\mu_{\mathrm{r}}$ with Gauss--Newton curvature. This provides greater
interior margin, avoids potentially indefinite Lagrangian curvature, and
allows contact-mode selection from a different basin. One final
Gauss--Newton solve at $\mu_{\mathrm{t}}$ tightens the restored trajectory.

The solve sequence terminates after the first successful phase and attempts at
most five finite solves. Success always requires the complete termination
conditions, including the unrelaxed physical residual
$r_{\mathrm{phys}}\leq\epsilon_{\mathrm{mpcc}}$. The recovery sequence is fixed
and problem-independent; objectives, constraints, and task tolerances are
unchanged.
	\section{Experimental Methodology}
	\label{sec:experiments}
	
\subsection{Benchmarks, Baselines, and Fairness}

The comparisons use benchmark formulations from the released CRISP
\cite{li2025crisp} and IMPACT \cite{li2026impact} source code. The CRISP set contains
Cartpole with Soft Walls, Push Box, Transport, and Push T; the IMPACT set
contains Push Box, Push T, and Cart Transport. For each comparison,
ContactIPM matches the baseline's physical parameters, horizon and
discretization, objective, constraints, initial state, target, and initial
guess. Both baselines are compiled and executed locally.

The robustness experiments use distinct problem instances, with each instance
solved once per method. The CRISP evaluation contains 15 Cartpole, 25 Push
Box, and 15 Transport initial-state/target combinations, together with all 50
Push-T cases from its released benchmark. Each IMPACT suite contains its
released source example and 49 predeclared local initial-state/target
variations. 

The timing experiments instead repeat one fixed instance per benchmark.
CRISP timing uses the released source examples and Push-T case
index~8, the ninth of the 50 released cases. IMPACT timing uses its
released source example for each benchmark. All
measurements were collected on an Intel(R) Core(TM) Ultra 7 155H
using release-mode executables and one software thread. After one warm-up,
each solver is run 20 times. In each repetition, ContactIPM and the baseline
run consecutively on logical processor~0, with their order randomized. Only
pairs for which both trajectories pass the common post-solve evaluation are
included. The closed-loop study uses 50
distinct disturbance rollouts comprising 1,104 MPC solves.

To separate OCP structure from complementarity treatment, we also transcribe
all seven CRISP- and IMPACT-parameter configurations in
acados~\cite{verschueren2022acados}. These transcriptions use the same problem parameters,
initialization, and task definitions as the corresponding released
benchmarks. Acados uses full SQP with partial condensing and HPIPM, but imposes
exact complementarity directly through $a\geq0$, $b\geq0$, and $ab\leq0$,
without an elastic relaxation.

	\subsection{Post-Solve Evaluation}
	\label{sec:post_solve}
	
	A solver return code alone is not counted as success. Each returned
	trajectory is processed by the same post-solve evaluator. A run is
	accepted only if the solver reports convergence, all trajectory values
	are finite, and the trajectory satisfies both the physical-feasibility
	and task-completion gates. Physical feasibility requires a maximum
	dynamics or algebraic-equality defect below $10^{-5}$, a maximum
	one-sided constraint violation below $10^{-8}$, and a maximum physical
	complementarity product below $10^{-5}$. The complementarity check
	evaluates
	\[
	\max_{k,p}\left|a_{k,p}b_{k,p}\right|
	\]
	using the original nonnegative contact quantities, rather than a
	solver's relaxed product row or elastic slack. Task completion is
	evaluated using the benchmark-specific terminal tolerances inherited
	from the corresponding released formulation.
	
	 Objectives and solution-quality
	metrics are likewise recomputed from the returned trajectories using
	the original benchmark costs. Solver convergence, physical feasibility,
	and task completion are therefore distinct diagnostics; a run contributes
	to the reported success count only when all three conditions hold.
	\section{Results}
	
	\label{sec:results}
		\begin{table}[t]
		\caption{Runtime Comparison ContactIPM vs CRISP on 20 Paired Runs}
		\label{tab:timing}
		\centering
		\resizebox{\columnwidth}{!}{
			\begin{tabular}{lccc}
				\hline
				Problem & ContactIPM [s] & CRISP [s] & Speedup [95\% CI]\\
				\hline
				Cartpole & 0.0272 & 0.2402 & 8.868 [8.417, 9.241]\\
				Push Box & 0.2614 & 0.5931 & 2.168 [2.128, 2.341]\\
				Transport & 0.1967 & 0.4988 & 2.337 [2.247, 2.595]\\
				Push T & 0.7885 & 5.5285 & 6.920 [6.669, 7.381]\\
				\hline
		\end{tabular}}
	\end{table}
	
	\begin{table}[t]
	     \caption{Robustness on benchmark suites from CRISP.}
		\label{tab:robustness}
		\centering
		\begin{tabular}{lcc}
			\hline
			Suite & ContactIPM & CRISP\\
			\hline
			Cartpole  & 15/15 & 15/15\\
			Push Box  & 24/25 & 19/25\\
			Transport & 8/15 & 8/15\\
			Push T & 50/50 & 27/50\\
			\hline
		\end{tabular}
	\end{table}
	
	\begin{table*}[t]
		\centering
		\caption{Median solution quality and feasibility on jointly successful
			CRISP-suite instances.}
		\label{tab:crisp_quality}
		\scriptsize
		\setlength{\tabcolsep}{2.4pt}
		\renewcommand{\arraystretch}{1.03}
		\begin{tabular}{lrrrrrrrr}
			\hline
			& \multicolumn{2}{c}{Cartpole}
			& \multicolumn{2}{c}{Push Box}
			& \multicolumn{2}{c}{Transport}
			& \multicolumn{2}{c}{Push T} \\
			\cline{2-9}
			Metric & ContactIPM & CRISP & ContactIPM & CRISP & ContactIPM & CRISP & ContactIPM & CRISP \\
			\hline
			Recomputed objective
			& 2.7202 & 2.7202
			& 18.6406 & 8.1969
			& 0.7921 & 0.7937
			& 6.7964 & 7.2313 \\
			
			Force Effort
			& 2.6213 & 2.6212
			& 18.5117 & 7.8774
			& 0.7225 & 0.7231
			& 2.9779 & 2.9558 \\
			
			Peak force
			& 10.113 & 10.115
			& 14.553 & 9.516
			& 8.276 & 8.356
			& 11.617 & 9.608 \\
			
			Terminal $e_p^N$
			& $2.38{\times}10^{-3}$ & $2.38{\times}10^{-3}$
			& $3.34{\times}10^{-2}$ & $4.43{\times}10^{-2}$
			& $4.96{\times}10^{-6}$ & $5.06{\times}10^{-6}$
			& $2.93{\times}10^{-3}$ & $2.89{\times}10^{-2}$ \\
			
			Terminal $e_{\theta/v}^N$
			& $2.98{\times}10^{-2}$ & $2.98{\times}10^{-2}$
			& $1.35{\times}10^{-2}$ & $2.13{\times}10^{-2}$
			& $6.96{\times}10^{-2}$ & $7.13{\times}10^{-2}$
			& $1.06{\times}10^{-4}$ & $8.55{\times}10^{-6}$ \\
			
			Max. equality defect
			& $4.44{\times}10^{-16}$ & $6.55{\times}10^{-8}$
			& $3.19{\times}10^{-16}$ & $1.40{\times}10^{-7}$
			& $1.70{\times}10^{-7}$ & $3.47{\times}10^{-15}$
			& $2.22{\times}10^{-16}$ & $1.84{\times}10^{-7}$ \\
			
			Max. complementarity residual
			& $9.94{\times}10^{-7}$ & $1.25{\times}10^{-14}$
			& $6.76{\times}10^{-7}$ & $6.14{\times}10^{-13}$
			& $9.41{\times}10^{-7}$ & $6.01{\times}10^{-14}$
			& $1.00{\times}10^{-6}$ & $9.26{\times}10^{-15}$ \\
			\hline
		\end{tabular}
		\vspace{1mm}
		\parbox{0.96\textwidth}{\scriptsize
		For Cartpole, Push Box and Push T,
		$e_p^N$ and $e_{\theta/v}^N$ denote terminal translation and angular
		errors; for Transport, they denote terminal position and velocity
		errors. Objectives use the original benchmark costs. Lower is better.}
	\end{table*}

	\subsection{Matched CRISP Comparison}
	
	Table~\ref{tab:timing} reports fixed-instance timing. All 80 timing repetitions pass the common post-solve evaluation. ContactIPM is faster on every problem, with paired
	median speedups from $2.168\times$ on Push Box to $8.868\times$ on Cartpole.
	The lower bound of every confidence interval is greater than
	one.
	
	Both methods produce physically valid trajectories on the timed instances.
	Cartpole solutions are nearly identical. On Transport, ContactIPM attains
	objective 2.965 versus 2.983 for CRISP. On Push T, it obtains lower tracking
	and effort components at the source weight. Push Box exposes a different local
	tradeoff: ContactIPM uses more force but reaches smaller terminal translation
	and rotation errors.
	
	Table~\ref{tab:robustness} reports robustness or generalization
	to genuinely different initial or target conditions. ContactIPM matches CRISP
	on Cartpole and Transport and succeeds on five additional Push Box cases. On
	these three initial/target suites, it passes 47/55 cases versus 42/55 for CRISP.
	
On Push T, CRISP reaches the terminal task gate in 43/50 cases, but
only 27 also satisfy the common physical-complementarity criterion. Reporting both values
	distinguishes failure to complete the task from a trajectory that appears
	successful in task space but violates the contact model.
	
	Table~\ref{tab:crisp_quality} compares solution quality and feasibility
	on the jointly successful cases from the multi-instance robustness
	suites in Table~\ref{tab:robustness}. Cartpole and Transport yield nearly identical objective and terminal-task
	quality. On Push Box, ContactIPM uses more contact effort and consequently has
	a larger objective, but reaches smaller terminal translation and angular
	errors. On the 27 jointly successful Push-T instances, ContactIPM obtains a
	lower objective and approximately tenfold smaller translation error, whereas
	CRISP uses slightly less effort and achieves a smaller angular error. CRISP
	drives complementarity closer to zero, while ContactIPM terminates near its
	prescribed $10^{-6}$ physical-complementarity tolerance.
	
	\begin{table}[t]
		\caption{Runtime Comparison ContactIPM vs IMPACT on 20 Paired Runs}
		\label{tab:impact_timing}
		\centering
		\resizebox{\columnwidth}{!}{
			\begin{tabular}{lccc}
				\hline
				Benchmark & ContactIPM [s] & IMPACT [s] & Speedup [95\% CI]\\
				\hline
				Push Box & 0.2666 & 0.0599 & 0.224 [0.217, 0.242]\\
				Push T & 0.2226 & 0.6645 & 2.960 [2.654, 3.201]\\
				Cart Transport & 0.0366 & 0.1799 & 4.908 [4.717, 5.255]\\
				\hline
		\end{tabular}}
	\end{table}
	
	\begin{table}[t]
		\caption{Robustness on benchmark suites from IMPACT.}
		\label{tab:impact_robustness}
		\centering
		\begin{tabular}{lcc}
			\hline
			Benchmark & ContactIPM & IMPACT\\
			\hline
			Push Box & 50/50 & 49/50\\
			Push T & 50/50 & 48/50\\
			Cart Transport & 50/50 & 50/50\\
			\hline
			Total & 150/150 & 147/150\\
			\hline
		\end{tabular}
	\end{table}

	\begin{table*}[t]
		\centering
		\caption{Median solution quality and feasibility on jointly successful
			IMPACT-suite instances.}
		\label{tab:impact_quality}
		\scriptsize
		\setlength{\tabcolsep}{5pt}
		\renewcommand{\arraystretch}{1.08}
		\begin{tabular}{@{}lrrrrrr@{}}
			\hline
			& \multicolumn{2}{c}{Push Box}
			& \multicolumn{2}{c}{Push T}
			& \multicolumn{2}{c}{Cart Transport} \\
			Metric
			& \multicolumn{1}{c}{ContactIPM}
			& \multicolumn{1}{c}{IMPACT}
			& \multicolumn{1}{c}{ContactIPM}
			& \multicolumn{1}{c}{IMPACT}
			& \multicolumn{1}{c}{ContactIPM}
			& \multicolumn{1}{c}{IMPACT} \\
			\hline
			Recomputed objective
			& 0.0116 & 0.0120
			& 0.0453 & 0.0384
			& $1.69\times10^{-4}$ & $2.72\times10^{-4}$ \\
			
			Force effort
			& 5.3512 & 3.7937
			& 1.4714 & 0.8690
			& 154.8426 & 254.0451 \\
			
			Peak force
			& 0.4443 & 0.4359
			& 0.4093 & 0.2447
			& 4.3689 & 7.0336 \\
			
			Terminal $e_p^N$
			& $3.63\times10^{-5}$ & $2.83\times10^{-5}$
			& $2.15\times10^{-4}$ & $5.34\times10^{-5}$
			& $5.45\times10^{-8}$ & $1.78\times10^{-5}$ \\
			
			Terminal $e_{\theta/v}^N$
			& $2.08\times10^{-5}$ & $1.62\times10^{-5}$
			& $7.66\times10^{-5}$ & $8.81\times10^{-5}$
			& $8.04\times10^{-9}$ & $9.92\times10^{-9}$ \\
			
			Max. equality defect
			& $2.78\times10^{-17}$ & $1.27\times10^{-8}$
			& $5.55\times10^{-17}$ & $3.91\times10^{-8}$
			& $2.22\times10^{-16}$ & $1.85\times10^{-8}$ \\
			
			Max. complementarity residual
			& $9.98\times10^{-7}$ & $3.15\times10^{-6}$
			& $1.00\times10^{-6}$ & $1.10\times10^{-6}$
			& $9.98\times10^{-7}$ & $2.93\times10^{-6}$ \\
			\hline
		\end{tabular}
		
		\vspace{1mm}
		\parbox{0.96\textwidth}{\scriptsize
			For Push Box and Push T,
			$e_p^N$ and $e_{\theta/v}^N$ denote terminal translation and angular
			errors; for Cart Transport, they denote terminal position and velocity
			errors. Objectives use the original benchmark costs. Lower is better.}
	\end{table*}
	\subsection{Matched IMPACT Comparison}

	Table~\ref{tab:impact_timing} reports fixed-instance timing.
	For each benchmark, both solvers execute the same released source case
	20 times. ContactIPM is $2.96\times$ faster on Push T and
	$4.91\times$ faster on Cart Transport, whereas IMPACT is
	$4.46\times$ faster on Push Box.

	Table~\ref{tab:impact_robustness} reports robustness on three
	50-case suites constructed from IMPACT's released benchmark
	formulations. Each suite contains the released source case and 49
	predeclared initial-state/target variations, with every instance solved
	once by each method. ContactIPM passes all 150 cases, whereas IMPACT
	passes 49/50 Push Box, 48/50 Push-T, and 50/50 Cart Transport cases,
	for 147/150 overall. The three differing outcomes give ContactIPM a small observed robustness advantage on these suites, with 150/150 accepted runs
	compared with 147/150 for IMPACT.
	
	Table~\ref{tab:impact_quality} reports median solution quality and
	feasibility over both solvers jointly successful cases from the multi-instance
	robustness suites. On Push Box, ContactIPM attains a slightly lower objective,
	whereas IMPACT uses less force and reaches slightly smaller terminal
	translation and angular errors. On Push T, IMPACT obtains a lower
	objective, force effort, peak force, and terminal translation error,
	while ContactIPM reaches a slightly smaller terminal angular error.
	
	On Cart Transport, ContactIPM obtains a lower objective, uses less
	force, and reaches smaller terminal position and velocity errors. Across
	all three suites, ContactIPM produces smaller median dynamics defects
	and physical-complementarity residuals.
	
	\begin{table*}[t]
		\centering
		\caption{Push Box robustness: ContactIPM vs Acados}
		\label{tab:acados_robustness}
		\scriptsize
		\setlength{\tabcolsep}{9pt}
		\renewcommand{\arraystretch}{1.05}
		\begin{tabular}{lccccc}
			\hline
			& ContactIPM & \multicolumn{4}{c}{Acados} \\
			Parameter set
			& Post-Solve Eval Accepted
			& Post-Solve Eval Accepted
			& Solver Success Exit
			& Physical Feasible
			& Goal reached \\
			\hline
			CRISP  & 24/25 & 0/25 & 19/25 & 25/25 & 0/25 \\
			IMPACT & 50/50 & 0/50 & 43/50 & 50/50 & 0/50 \\
			\hline
		\end{tabular}
	\end{table*}
	
	\subsection{OCP Structure without Elastic Complementarity}
	\label{sec:acados_baseline}
	
	To investigate whether exploiting OCP structure alone is
	sufficient for contact-implicit optimization, we transcribe all
	shared benchmarks in acados~\cite{verschueren2022acados}. Acados exploits the
	multistage structure through full SQP, partial condensing, and HPIPM, but does
	not use an elastic complementarity treatment. Instead, it directly imposes
	$a\geq0$, $b\geq0$, and $ab\leq0$. Because both factors are nonnegative, this
	enforces exact complementarity rather than a relaxation. All seven
	formulations build and execute successfully.
	
	A run is accepted only if: (i) the solver reports successful convergence,
	(ii) an independent post-solve evaluation verifies the dynamics, side constraints,
	algebraic equalities, and physical complementarity, and (iii) the trajectory
	reaches the terminal task goal. These conditions are also reported separately
	for acados to distinguish solver failure, physical infeasibility, and failure
	to complete the task.

	Table~\ref{tab:acados_robustness} exposes the contact-initiation failure most
	clearly. Acados passes the physical feasibility check in all 75 Push Box cases and reports
	successful convergence in 62, but none reaches the terminal goal. Inspection
	shows that it frequently terminates at a physically feasible zero-contact
	trajectory in which the box does not move. Thus, neither solver status nor
	physical feasibility alone establishes a successful contact plan.
	
	The same trend appears on the other contact-rich benchmarks. To show that the difficulty is not specific to Push Box, we next consider the remaining contact-rich manipulation suites: CRISP Transport, CRISP Push T, IMPACT Push T, and IMPACT Cart Transport. Across these 165 cases, acados produces only one accepted trajectory, whereas ContactIPM succeeds in 158. OCP
	structure therefore addresses the temporal linear algebra but does not remove
	the degeneracy of exact complementarity. In particular, the exact feasible set
	has no interior region in which both complementary quantities are positive,
	and the product derivative vanishes at a biactive point. A local SQP method
	can consequently remain in a zero-contact basin or stall while changing
	contact modes.
	
	ContactIPM combines the missing MPCC treatment with OCP-structured linear
	algebra. Its elastic lift provides an interior path for initiating and changing
	contact, while stagewise elimination preserves the Riccati structure. The
	relaxation is tightened during optimization, and final acceptance still
	requires the original physical complementarity residual. Because acados and ContactIPM also differ in globalization and
	Newton-system construction, this comparison is not a one-factor
	ablation. Nevertheless, the results show that exploiting OCP structure
	alone does not provide robust MPCC performance from the tested
	initializations.
	\subsection{Closed-Loop Validation}
	\label{sec:closed_loop_validation}
	
	We evaluate ContactIPM in closed-loop control using the Push Box
	quasi-static model. At each 0.1\,s control update, the controller observes the
	box pose, solves a 19-stage target-tracking OCP, and applies the first control.
	The first solve uses zero free-state and control guesses, while subsequent
	solves use shifted warm starts. The simulated plant is integrated using ten
	RK4 substeps per control period. A rollout is successful when both the
	translational and angular target errors remain below 0.1\,m and 0.1\,rad,
	respectively, for ten consecutive control steps.
	
\begin{figure*}[t]
		\centering
		\includegraphics[width=0.85\textwidth]
		{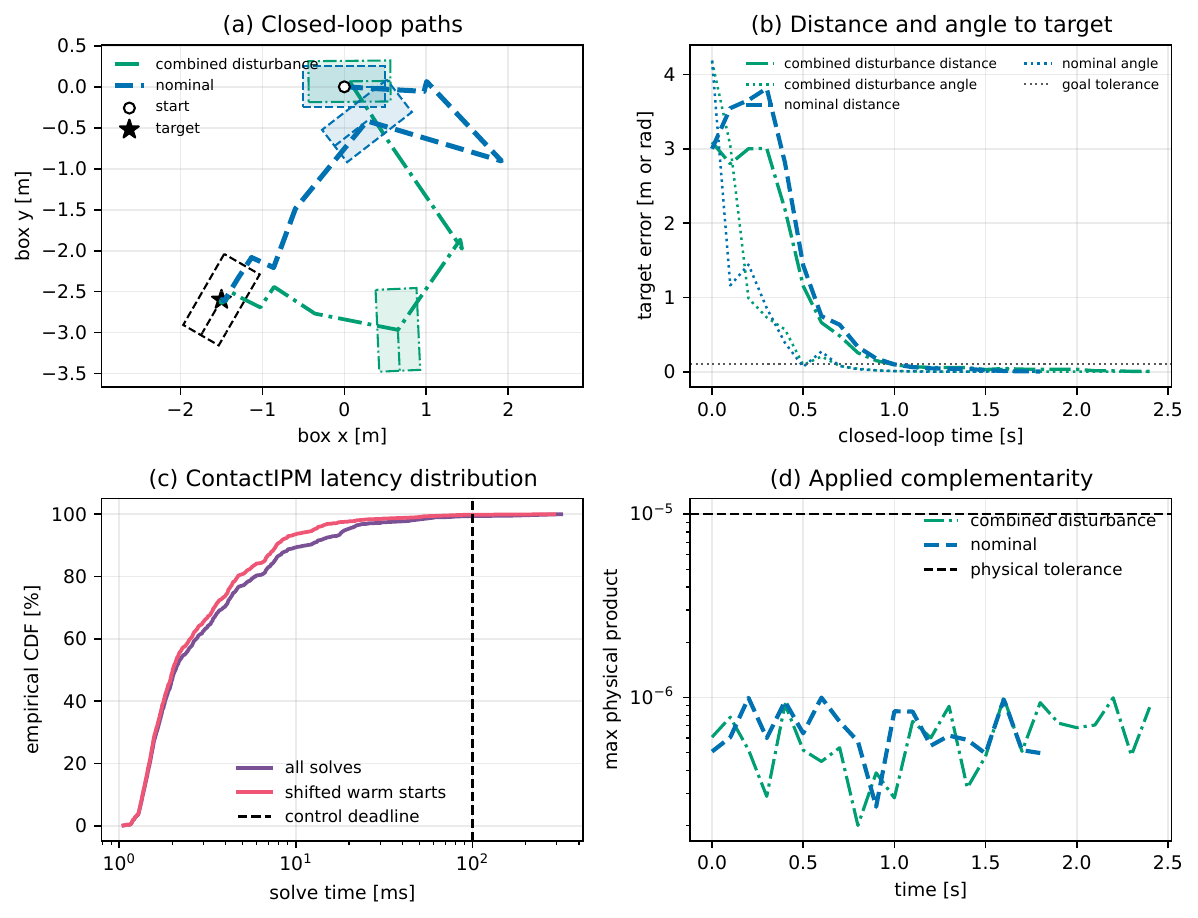}
		\caption{
			Closed-loop Push Box validation.
			(a) Representative nominal and combined-disturbance trajectories.
			Shaded footprints show the initial and geometric-halfway box poses,
			while the unfilled dashed footprint denotes the target. Each
			center-to-front segment indicates the signed box orientation.
			(b) Translational and angular target errors versus elapsed closed-loop
			time. Samples are separated by the 0.1\,s control period; the horizontal
			dotted line is the common 0.1\,m or 0.1\,rad goal tolerance. Curves terminate
			after both tolerances have been maintained for ten consecutive steps.
			(c) Empirical distribution of solve times over all 1,104 MPC solves;
			the vertical line denotes the 100\,ms control deadline.
			(d) Original physical complementarity products evaluated independently
			of the elastic product slack and its logarithmic barrier.
		}
		\label{fig:closed_loop_validation}
	\end{figure*}

	We consider 50 rollouts distributed across the following conditions: nominal motion, initial-pose
	perturbations, mass and friction mismatch, isolated pose-state resets, and a
	combined-disturbance scenario. The combined disturbance includes an initial
	pose error, independently perturbed mass and friction, measurement noise at
	every feedback update, and one scheduled pose-state reset. Representative nominal and combined-disturbance closed-loop
	trajectories are shown in Fig.~\ref{fig:closed_loop_validation}(a).
	
	Table~\ref{tab:closed_loop_robustness} reports the distribution of closed-loop
	completion time and final pose accuracy. A rollout is successful only after
	both the position and angular errors remain below 0.1\,m and 0.1\,rad,
	respectively, for ten consecutive control steps. The step columns count the
	0.1\,s feedback updates required to complete this goal hold. The median
	describes a typical rollout in each scenario, while the maximum reports the
	worst observed rollout. A fractional median, such as 20.5 steps, results from
	averaging the two central values of an even-sized group; no individual
	rollout uses a fractional control step. All 50 rollouts succeed. Nominal motion completes in 19 steps, corresponding
	to approximately 1.9\,s. The combined disturbance increases the median
	completion time to 25 steps and the maximum to 27 steps, but its worst final
	position and angular errors are only 11.9\,mm and 2.15\,mrad. Thus, the tested
	disturbances increase completion time and final-error variability without
	causing task or solver failure.
	
	\begin{table*}[t]
		\centering
		\caption{Closed-loop Push Box robustness.}
		\label{tab:closed_loop_robustness}
		\scriptsize
		\setlength{\tabcolsep}{7pt}
		\renewcommand{\arraystretch}{1.08}
		\begin{tabular}{lcrrrrrr}
			\hline
			& & \multicolumn{2}{c}{Control steps}
			& \multicolumn{2}{c}{Final position [mm]}
			& \multicolumn{2}{c}{Final angle [mrad]} \\
			Scenario & Success
			& Median & Maximum
			& Median & Maximum
			& Median & Maximum \\
			\hline
			Nominal
			& 5/5
			& 19 & 19
			& 0.153 & 0.153
			& 0.161 & 0.161 \\
			
			Initial-pose perturbation
			& 15/15
			& 19 & 21
			& 1.57 & 10.2
			& 0.0868 & 0.876 \\
			
			Mass/friction mismatch
			& 10/10
			& 20.5 & 24
			& 0.523 & 2.79
			& 0.0973 & 0.969 \\
			
			Isolated pose reset
			& 10/10
			& 26 & 31
			& 0.361 & 7.32
			& 0.0800 & 0.801 \\
			
			Combined disturbance
			& 10/10
			& 25 & 27
			& 1.36 & 11.9
			& 0.621 & 2.15 \\
			\hline
		\end{tabular}
	\end{table*}
	
	As shown in Fig.~\ref{fig:closed_loop_validation}(b), the pose errors
	generally decrease because the measured state is repeatedly incorporated into
	the target-tracking MPC problem. The trajectories have different durations
	because each controller run terminates only after satisfying the ten-step goal
	hold.
	
	Across all 1,104 solves, the median, 90th-percentile, and 99th-percentile solve
	times are 2.08\,ms, 11.5\,ms, and 60.8\,ms, respectively. The solve time CDF is shown in Fig.~\ref{fig:closed_loop_validation}(c). Seven solves exceed
	the 100\,ms control period. Among the 1,054 shifted-warm-start solves, the
	median time is 2.01\,ms and only two solves  ($0.19\%$) miss the 100\,ms deadline.
	The controller meets the 100\,ms deadline on 1097/1104 solves
	($99.37\%$) overall and on 1052/1054 shifted-warm-start solves
	($99.81\%$).
	
	The post-solve evaluation gives maximum planned dynamics and side-constraint
	violations of \(8.62\times10^{-7}\) and \(7.30\times10^{-9}\), respectively. The time evolution of physical complementarity is shown in Fig.~\ref{fig:closed_loop_validation}(d).
	The largest planned and applied physical complementarity products are
	\(1.01\times10^{-6}\) and \(1.00\times10^{-6}\). Thus, closed-loop feedback
	does not compromise the primal feasibility or physical complementarity of the
	computed plans.
	
	Finally, because the quasi-static benchmark contains pose but no momentum
	state, the scheduled reset is implemented as a direct change in position and
	yaw rather than as an inertial impulse. This experiment therefore evaluates
	closed-loop recovery and complementarity handling under model mismatch and
	state disturbances.

	\section{Conclusion}
	\label{sec:conclusion}
	
	ContactIPM demonstrates that primal--dual interior point methods need not be
	abandoned for contact-implicit trajectory problems. By combining explicit
	complementarity recognition, elastic interior relaxation, stagewise
	primal--dual elimination, Riccati recursion, scaling, and multi-phase recovery, the
	solver targets the gap between MPCC-specific robustness and
	structure-exploiting optimal-control optimization. It is faster than CRISP on
	all four timed instances and improves aggregate success on the
	multi-instance suites. Against locally executed IMPACT, it
	passes all 150 cases and is faster on Push T and Cart Transport, while
	IMPACT is substantially faster on Push Box. Across these heterogeneous benchmarks, ContactIPM establishes that MPCC reliability, primal--dual consistency, and OCP-structured computation can be realized together in a single practical solver.
	

\end{document}